# Object-Agnostic Suction Grasp Affordance Detection in Dense Cluster Using Self-Supervised Learning

Mingshuo Han, Wenhai Liu., Zhenyu Pan, Teng Xue, Quanquan Shao, Jin Ma, Weiming Wang,
Institute of Knowledge Based Engineering, Shanghai Jiao Tong University

*Abstract*—In this paper we study grasp problem in dense cluster, a challenging task in warehouse logistics scenario. By introducing a two-step robust suction affordance detection method, we focus on using vacuum suction pad to clear up a box filled with seen and unseen objects. Two CNN based neural networks are proposed. A Fast Region Estimation Network (FRE-Net) predicts which region contains pickable objects, and a Suction Grasp Point Affordance network (SGPA-Net) determines which point in that region is pickable. So as to enable such two networks, we design a self-supervised learning pipeline to accumulate data, train and test the performance of our method. In both virtual and real environment, within 1500 picks (~5 hours), we reach a picking accuracy of 95% for known objects and 90% for unseen objects with similar geometry features.

## I. INTRODUCTION

Suction grasping is widely used for "pick-and-place" tasks in domain of industry and warehouse logistics. In warehouse order fulfillment scenario, objects are randomly stacked in boxes. It results in difficulties when executing grasp of object due to abnormal pose, unseen object, occlusion etc. Compared to multifingered hand-like end-effector, vacuum suction cup has advantage owing to its higher efficiency and reaching ability in complex environment, while relatively fewer researches have been extended on suction grasping [1].

Traditional solution for suction grasping searches for planar part calculated by point cloud near the centroid of object [2] [3] [4] to execute the grasp, which relies highly on the precision of the seal model of suction pad. When facing with complex tasks, deep learning aided vision methods are gradually becoming solution [5] [6], which benefit from human-labeled data. Self-supervised learning [7], reinforcement learning [8] are two other ways to acquire a grasp position and pose or a series of control policy for grasping. However, these researches are still mainly concentrated on multifingered hand-like end-effector and the experiment settings in these researches are still simplified.

In this work, we propose a vision-based two-step method to detect object-agnostic suction affordance in dense cluster (See Figure 1). In suction grasp procedure, firstly, the suction region estimation network (FRE-Net) acquires a region which has the highest suction grasp probability from several proposals. Then in this region, the suction point grasp affordance network (SGPA-Net) predicts which point could be a successful suction grasp. Within a greedy policy self-supervised learning procedure, these two networks are trained and tested (See Figure 4).

Mingshuo Han, Wenhai Liu., Zhenyu Pan, Teng Xue, Quanquan Shao, Jin Ma, Weiming Wang are with the Institute of Knowledge Based Engineering, School of Mechanical Engineering, Shanghai Jiao Tong University, Shanghai 200240, China. (corresponding author phone: 008621-34208531; fax: 008621-34506820; e-mail: wangweiming@sjtu.edu.cn, {mrjay101, sjtu-wenhai, pan13855965084,xueteng, shao124,heltonma}@sjtu.edu.cn).

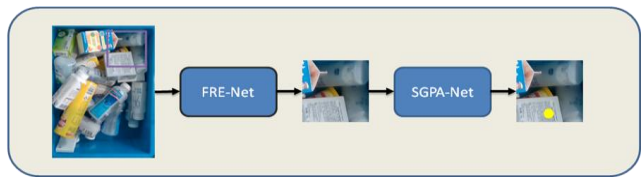

Figure 1. Two-step suction grasp affordance detection

The main contributions of this paper are:

- Two networks with similar structure are introduced, one for fast region estimation, the other for suction point affordance detection. These two networks ensure object-agnostic suction grasp with high speed and little omission ratio.

- A full-automatic self-supervised learning pipeline with greedy policy is proposed for training both networks in sequence.

- To verify the learning pipeline and the networks performance, both virtual and real experiment environments are setup. With the processing of experiment, a database containing region estimation data and suction grasp point affordance data from both virtual and real test is created.

The rest of the paper is organized as follows. Related work is reviewed in Section 2. Picking configuration, network structures, self-supervised learning pipeline and the accumulation of data are presented in Section 3. The evaluation of the method in both real and virtual environment is discussed in Section 4. Finally, in Section 5, the conclusion of this paper and prospect for future work.

## II. RELATED WORK

In this section, we review work related to our main methods in recent years. Researches on suction grasping and object grasp learning are discussed.

### A. Suction grasping

During the three times Amazon Picking Challenge in sequence, where competitors proposed solution to "pick-and-place" tasks for warehouse logistics scenario, manipulators of suction vacuum cup were widely applied. In APC 2015, more than half of teams integrated suction cup into their system [9]. And Eppner et al. [10] won by pushing the object against the floors, walls or other objects to achieve suction grasp. In 2016, Team Delft [11] won by executing suction grasp on the planar part near the centroid of objects, with the aid of 6-D pose estimation. And Team NimbRo, who achieved second and third place, executed grasp on points determined heuristically or from 6D item model registration. In 2017, Team MIT-Princeton [12] mapped the vision input to probability of suction grasp by fully convolutional networks. But it should be noted that these solutions are target-guided, which means they are specifically designed for specific objects.

Besides the competition, Domae et al [13] estimated graspability on a single depth map by representing two mask image – one describing contact region, the other describing collision region. Mahler et al. [1] generated a dataset called Dex-net 3.0 with 2.8 million point clouds, suction grasps, and grasp robustness labels. Then they trained a Grasp Quality Convolutional Neural Networks with Dex-Net 3.0, which could classify robust suction targets in point clouds. Liu et al. [5] proposed a double stream CNN to predict point suction affordance in dense cluster with human-labeled data. In summary, the non deep-learning methods' generalization ability is not sufficient, while the deep-learning methods on suction grasp are still trained on human labeled data.

### B. Robot grasp learning

Human-labeling, self-supervised learning and reinforcement learning are 3 major methods for robots to learn to grasp object in a specific task.

The deep learning methods mentioned in last paragraph could be classified into human labeling [1] [5]. As far as we know, Lenz et al [15], who was the first trying to solve the grasp problem with neural networks, proposed a grasp cascade with two neural networks, one pruning out unlikely candidate grasp, the other predicting the most likely grasp pose. Furthermore, with data labeled by human or adversarially generated, researchers have also studied problems like pose estimation [16] [17] [18], task learning [19] for robot grasp manipulation.

If a grasp is regarded as a single action determined by end-effector's pose, then self-supervised learning enables a robot itself to accumulate data and extract feature for grasp. With a dataset size of 50K data points collected over 700 hours of robot grasping attempts, Pinto et al [7] trained a CNN to predict grasp angle and grasp center from random candidates. By the difference caused by interactions between robot and environments, Devin et al [20] convert object visual features to vector by CNN. And R.Florence et al [21] trained neural networks as object feature descriptors by robot manipulation. Fang et al [22] proposed a network with two modules to jointly optimize both task-oriented grasping of a tool and the manipulation policy for that tool by self-supervision in simulation.

Besides, reinforcement learning considers a grasp is a Markov Decision Process. Using a self-supervised vision-based reinforcement learning framework, Kalashniko et al [8] enabled robot to continuously optimized its grasp strategy. With 580k real-world grasp, the grasp success rate for unseen object reached 96%. Quillen et al [23] did a simulated comparative evaluation of off-policy reinforcement learning method for robotic grasping. Zeng et al [24] studied especially synergies between pushing and grasping with deep Q-learning, reaching 83% grasp success rate.

It is easy to find that most of the learning methods are for multifingered end-effector. But what is hard to catch is that, suction grasp calls for more accurate affordance prediction, cause the tolerance for a suction grasp is much smaller than a hand-like grasp. We study the grasp problem with vacuum suction pad, with vision-based deep-learning detection method, and also a learning-pipeline to enable robot automatically learn to converge this method.

### III. PROBLEM FORMULATION AND METHOD

Our study aims at, with vacuum suction cup, picking out every object randomly stocked in box at different pose. High success rate, low omission ratio and high speed are the main targets of our method.

#### A. Suction grasp configuration

***Suction grasp point affordance:*** In a (100, 100) pixels region, we extract 289 (17, 17) suction grasp point candidates. The suction grasp affordance feature of one point is described by the RGB image and depth image with size of (32,32) pixels around it. With the 289 pairs of RGB image and depth image as input, the SGPA-Net classifies them as 0 (a bad grasp) or 1 (a good grasp). (See Figure 2)

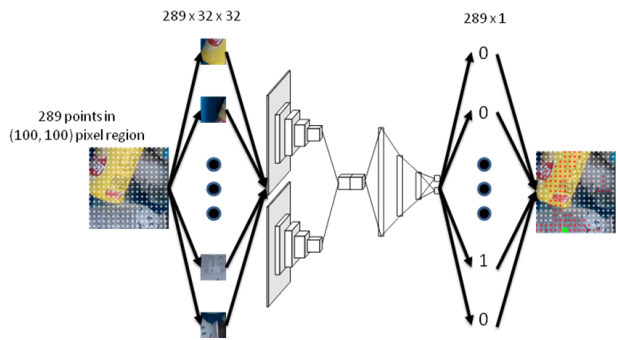

Figure 2. Suction grasp point affordance detection

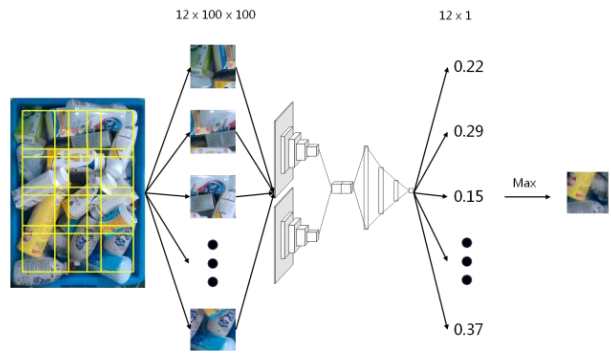

Figure 3. Fast region estimation

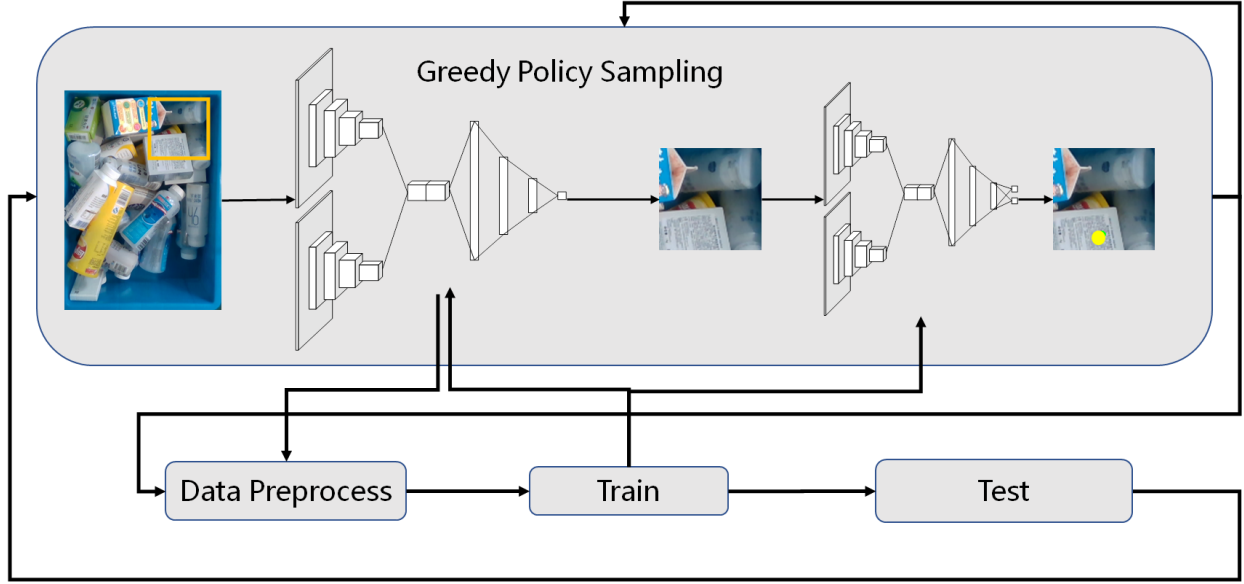

Figure 4. Self-supervised learning pipeline for two-step suction grasp affodance detection

*Suction region estimation:* Before determining which point could be a good grasp, the robot should figure out which regions contain pickable objects. We extract region candidates by sliding window with (100, 100) pixels but with large stride. 12 region candidates are drew from a box with nearly (200, 300) pixels. Corresponding to these 12 candidates, a set of suction score would by generated by FRE-Net's regression layer. With the highest suction score, the region contains pickable objects is selected. (See Figure 3)

### B. SGPA-Net and FRE-Net

The suction point affordance detection network and region estimation network have similar structure, differing on the fully connected layers. The input is RGB image patch and (depth, depth, depth) image patch. VGG-16 net's convolutional layers [25] pre-trained on ImageNet are used to extract 2-D features, followed by one convolutional layer to integrate them. Then there are several fully connected layers lead to output layer - a binary classification for SGPA-Net or a regression for FRE-Net. (See Figure 2 and Figure 3)

### C. Self-supervised learning pipeline

To train the two networks mentioned above, a self-supervised learning pipeline combined with the training and test procedure is introduced (See Figure 4). Greedy policy sampling is executed continuously. Every 100 picks Data Preprocess, Train and Test part manipulates on data, trains networks and tests the networks' performance.

*Greedy policy sampling:* During the training, in order to balance quantity of negative and positive sample, some percentage of the sampling should be executed by greedy policy. That means executing a suction grasp on points generated by temporary networks. This exponential formula defines the percentage of greedy pick, in which $\alpha_E=0.05$, $\alpha_S=0.8$, $\alpha_D=2000$, $e=2.71828$ and n is the iteration step.

$$p = 1 - (\alpha_E + (\alpha_s - \alpha_E)) * e^{-\frac{n}{\alpha_D}} \quad (1)$$

*Pick and place:* One time "pick-and-place" consists of five phases: "estimate possible region", "figure out picking point", "executing picking", "obtain feedback signal from sensor", "place picked object", and "log and image recorded". Due to the complexity of environment, the movement of the suction pad is limited to "up" and "down", which simplifies motion planning.

*Dataset process and database creation:*

SGPA-Net: We save, for one suction grasp point, RGB image, depth image both with (32, 32) pixels, and the corresponding sensor feedback signal (0 or 1). During accumulating data for SGPA-Net, we notice that the vacuum suction pad performance is unstable when facing some circumstances, like edge of object or surface with non-vertical normal vector. To make the networks converge fast, depth images with large gradient are all relabeled as negative sample. Besides, images are rotated 16 times to enlarge the dataset. (See Figure 5)

FRE-Net: Only a binary value is not sufficient for describing a region contains how many possible good points. Thus, we make use of latest SGPA-Net to generate the label for FRE-Net. On one (100, 100) pixel region, a (17, 17) binary map is generated by SGPA-Net with every point's surrounding RGB image patch and (depth, depth, depth) image patch as input. Then this region is labeled by the percentage of the good grasp point on that binary map.

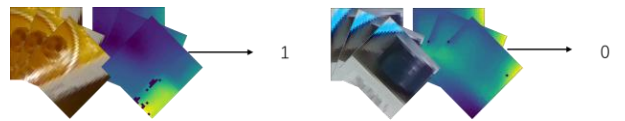

Figure 5. SGPA-Net dataset creation

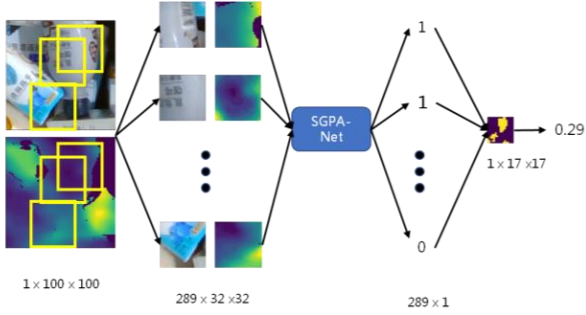

Figure 6. FRE-Net dataset creation

*Train:* Every 100 picks we do training process. On a GTX-1080 Ti GPU, within Pytorch deep learning framework we train both the two networks. 70% of the dataset is used as training set, and 30% as validation set. Parameters for the training process are listed below.

TABLE I.  NETWORKS TRAINING PARAMETERS

|  | Networks training parameters | |
| --- | --- | --- |
|  | *SGPA-Net* | *FRE-Net* |
| Optimizer | SGD | Adam |
| Loss fuction | CrossEntropyLoss | MSELoss |
| Learning rate | 0.001 | 0.001 |
| momentum | 0.9 | - |
| Weight decay | $10^{-5}$ | - |
| Training Epoch | 100 | 300 |

*Test:* Test is done with the trained network to verify if the networks work well. Firstly, FRE-Net chooses a region with highest percentage of pickable points from 12 region candidates. Then from the chosen region, SGPA-Net predicts suction grasp affordance among 289 (17, 17) candidates. However, through the SGPA-Net, these 289 candidates generate 289 binary values, which could be reformed into a binary map with size of (17,17). We still have to determine which is the best point. A convolutional kernel E [5] to optimize the suction point selection is proposed. Performing convolutional operation on the binary map generated by SGPA-Net, we could acquire the most centered pickable point with the highest value. (See Figure 7)

$$E=\begin{bmatrix} 0.1 & 0.3 & 0.5 & 0.3 & 0.1 \\ 0.3 & 0.5 & 0.8 & 0.5 & 0.3 \\ 0.5 & 0.8 & 1 & 0.8 & 0.5 \\ 0.3 & 0.5 & 0.8 & 0.5 & 0.3 \\ 0.1 & 0.5 & 0.8 & 0.3 & 0.1 \end{bmatrix} \quad (2)$$

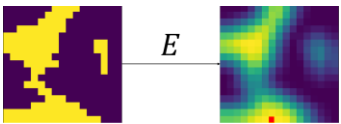

Figure 7. FRE-Net dataset creation

## IV. EXPERIMENTS AND RESULTS

To test whether our networks and pipeline work, we implement test in both simulation and real-world environment.

### A. Implemention in simulation

We use V-REP as simulation tool. A suction cup made of force connector is attached at the end joint of a UR5 robot arm. The force connector is also used as a sensor which detects whether an object is picked up. 10 kinds of objects' mesh models from YCB-object are randomly scattered in a box with 50cm length, 30cm width, 15cm depth. An RGB-D camera is mounted on the top of the box, where it could cover the whole box. (See Figure8). Every 50 picks, objects are randomly rearranged.

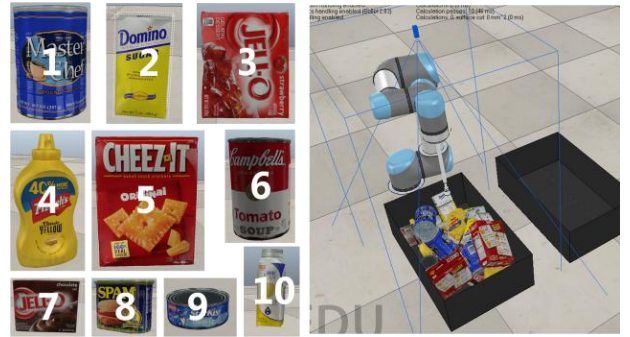

Figure 8. Simulation implemention

### B. Implemention in real-world environment

In reality, our system is composed of a UR5 robot arm, RealSense D435 RGB-D camera and a pneumatic vacuum suction pad and a SMC vacuum switch as sensor. The objects are 15 kinds of daily packages in our life, in which 5 kinds of them aren't used in training but for unseen object test. One point differs most from the virtual environment is that we couldn't rearrange automatically objects in box during test. Thus, we use 2 boxes as container, picking random times from box A to box B, and then doing the same from box B to box A.

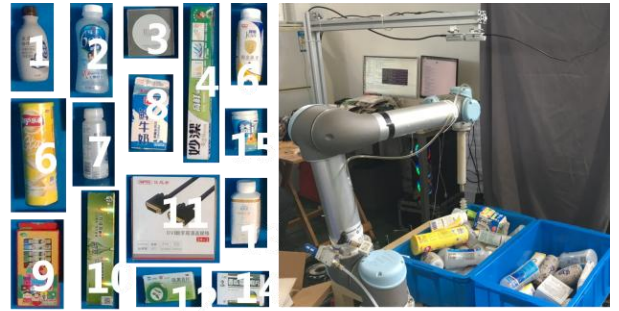

Figure 9. Real-world implemention

### C. Result

In simulation, our robot did 1000 times "pick-and-place", while in real-world 1500 times are done. After every 100 picks, dataset preprocess, training of the networks are done. After every training procedure, we test networks at that time for 50 times to verify the performance. With train and test added, 1500 times real-world suction grasp learning procedure takes only 5 hours. And at the end of the both virtual and real-world learning procedure, we also did 150 times "pick-and-place" test for the latest networks, which should have the best performance. The results for suction grasp success rate are shown below (See Figure 10, 11 and 12). We also demonstrate our method's advantage by comparision (See Figure 13).

Another amazing result is that the FRE-Net's omission rate is really low，with a relatively high detection speed. To prove that, we test other region selection methods like a fully convolutional network (FCN) trained with human-labeled data and using SGPA-Net compute all the points contained in box. The results are listed in Table Ⅱ, in which omission rate means the ratio that a region without object is chosen.

TABLE II. REGION DETECTION METHOD

|  | Region Detection Methods | | |
| --- | --- | --- | --- |
|  | **FRE-Net** | **FCN** | **Fully Covered SGPA-Net** |
| Time | 0.5s | 0.4s | 6s |
| Omission rate | 2% | 15% | ≈ 0 |

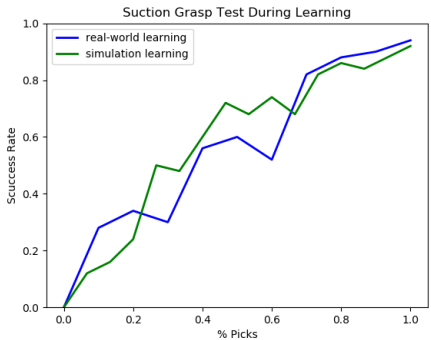

Figure 10. Suction grasp test during learning: Although 50 times test is not enough to reveal the true performance of temporal networks,but the rissing tendance of success rate is still obvious. Both the success rate rise along with more picks. We stopped the self-supervised learning pipeline when the success rate is over 90%, to be exact, 92% for simulation after 1000 picks and 94% for real-world after 1500 picks.

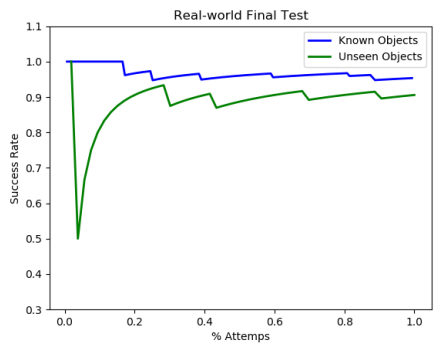

Figure 11. Real-world final test: In real-world implemention, we did seperately 150 times suction grasp test to verify the performance of latest networks with 10 objects during training and 5 unseen objects. As is shown above, the suction grasp success rate reaches 95% for known objects and 90 for unseen objects.

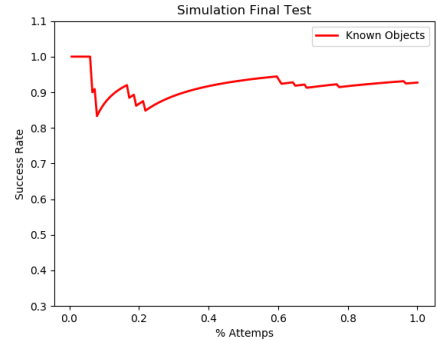

Figure 12. Simulation final test: For simulation implemetion we only tested the performance of latest networks for known objects. As is shown, 92.5% suction grasp success rate is reached.

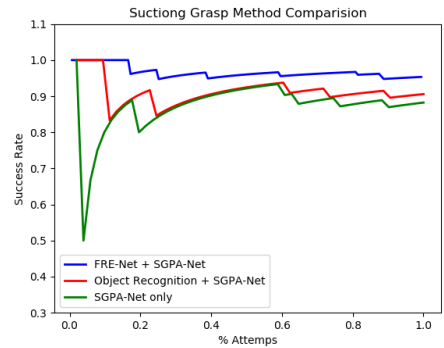

Figure 13. Suction grasp method comparision: We test 3 suction affordance detection methods –As we can see above, a)FRE-Net + SGPA-Net performs best on success rate. b) For lack of ability to recongnize unseen objects, the Object Recongnition + SGPA-Net only reachs 91%. c) Randomly generated region +SGPA-Net detection is the worst, which failes a lot when few objects left in box.

## V. CONCLUSION

This paper presents a two-step object-agnostic suction affordance detection method for suction grasp in warehouse logistics and its self-supervised learning pipeline. Making use of RGB image and depth image, Fast Region Estimation Network (FRE-Net) predicts which region contains pickable objects. Then, Suction Grasp Point Affordance Network (SGPA-Net) detects which point in the chosen region is pickable. The whole detection takes less than 1 second. The SGPA-Net is also used for generating label for FRE-Net during data process. In simulation and real-world experiments, within less than 1500 times "pick-and-place", we achieve 95% suction grasp success rate for known objects and 90% for unseen objects. Greedy policy sampling, dataset processing and a convolutional kernel optimizing suction point selection all contribute to the training process and test accuracy.

In future work, at first place, we urge to combine two networks together and train them with mixed loss. Moreover, on the one hand, in order to apply directly the networks trained in virtual environment, a transfer learning method will be researched to get over the gap between virtual and reality. On the other hand, we want to eliminate human's instinct judgement as much as possible, so a suction grasp feature representation which guides the sampling process is in to-do list.

ACKNOWLEDGMENT

This research is supported by Special Program for Innovation Method of the Ministry of Science and Technology, China (2018IM020100), and National Natural Science Foundation of China (51775332, 51675329, 51675342).